\documentclass[10pt,twocolumn,letterpaper]{article}

\usepackage[pagenumbers]{wacv}              %

\usepackage[breaklinks,colorlinks]{hyperref}

\newif\ifshowcomments								
\showcommentstrue				%

\usepackage{booktabs}
\usepackage{wrapfig}

\usepackage[all]{nowidow} %
\usepackage{amsmath}
\usepackage{amsfonts}
\usepackage{amssymb}
\usepackage[pdftex]{graphicx}
\graphicspath{{graphics/}}
\usepackage{trfsigns}
\usepackage{caption} %
\usepackage{subcaption} %
\usepackage{enumitem}
\usepackage{color}
\usepackage{calc}
\usepackage{tabularx}
\usepackage{booktabs}
\usepackage{ragged2e}
\usepackage{hyperref}

\usepackage[acronym]{glossaries}
\usepackage{siunitx}

\usepackage{xstring}
\usepackage{catchfile}

\usepackage{adjustbox}
\newcommand{\lastretrieveddate}{2023-03-29}

\ifshowcomments
    \newcommand{\margtodo}
    {\marginpar{\textbf{\textcolor{blue}{ToDo}}}{}}
    \newcommand{\todo}[1]
    {{\textbf{\textcolor{blue}{(\margtodo{}#1)}}}{}}
    
    \newcommand{\margmajortodo}
    {\marginpar{\textbf{\textcolor{red}{ToDo}}}{}}
    \newcommand{\imp}[1]
    {{\textbf{\textcolor{red}{(\margmajortodo{}#1)}}}{}}
    
    \newcommand{\margidea}
    {\marginpar{\textbf{\textcolor{green}{Check}}}{}}
    \newcommand{\tocheck}[1]
    {{\textbf{\textcolor{green}{(\margidea{}#1)}}}{}}
    
    \newcommand{\margtocheck}
    {\marginpar{\textbf{\textcolor{cyan}{Idea}}}{}}
    \newcommand{\idea}[1]
    {{\textbf{\textcolor{cyan}{(\margtocheck{}#1)}}}{}}
    
    \newcommand{\margtochange}
    {\marginpar{\textbf{\textcolor{cyan}{Change}}}{}}
    \newcommand{\tochange}[1]
    {{\textbf{\textcolor{cyan}{(\margtochange{}#1)}}}{}}
\else
    \newcommand{\todo}[1]{}
    \newcommand{\imp}[1]{}
    \newcommand{\idea}[1]{}
    \newcommand{\tocheck}[1]{}
    \newcommand{\tochange}[1]{}
    
\fi
\definecolor{mygrey}{gray}{0.65}
\ifshowcomments  %
    
    \newcommand{\pcite}[2]{{\color{gray}(\cite{#1} #2)}}  %
    \newcommand{\margcomment}
    {\marginpar{\textbf{\textcolor{mygrey}{Com.}}}{}}
    \newcommand{\scomment}[1]
    {{\textit{\textcolor{mygrey}{(\margcomment{}#1)}}}{}}
\else  %
    \usepackage{comment}
    \newcommand{\pcite}[2]{}
    \newcommand{\scomment}[1]{}
    
\fi
\usepackage[style=ieee,  %
    doi=false,
    url=true,
    isbn=false,
    mincitenames=1,
    maxcitenames=1,
    minbibnames=6,
    maxbibnames=6,
    backend=biber]{biblatex}  %

\DeclareSourcemap{
    \maps[datatype=bibtex, overwrite]{
        \map{
            \step[fieldset=note, null]
            \step[fieldset=urldate, null]
        }
        \map[overwrite=true]{ %
            \pertype{online}  %
            \step[fieldset=urldate,fieldvalue={Last accessed \lastretrieveddate{}}]
        }
    }
}

\renewcommand{\wrt}{w.r.t.\ } %

\newcommand{\idest}{i.e.\ } %
\renewcommand{\eg}{e.g.\ } %
\renewcommand{\cf}{cf.\ } %

\newcommand{\maskrcnn}{Mask {R-CNN}}
\newcommand{\cubercnn}{Cube {R-CNN}}

\newcommand{\resnetfpn}{ResNet-50-FPN}

\newcommand{\dstamp}{TAMPAR}

\newcommand{\dsparcel}{Parcel3D}

\newcommand{\dsreal}{Parcel2D Real}

\newcommand{\ap}[1]{$\text{AP}_{#1}$}
\newcommand{\boxap}[1]{$\text{Box AP}_{#1}$}
\newcommand{\maskap}[1]{$\text{Mask AP}_{#1}$}

\newcommand{\kpap}[1]{$\text{Keypoint AP}_{#1}$}

\newcommand{\urltampar}{\href{https://a-nau.github.io/tampar}{https://a-nau.github.io/tampar}}

\newcommand{\kpi}[1]{$k_{#1}$}
\newcommand{\vuss}[2]{(#1, #2)}

\newglossaryentry{i4.0}{name={Industry 4.0}, description={Industry 4.0}}
\newglossaryentry{forklift}{name={forklift truck}, description={}, plural={forklift trucks}}
\newacronym[longplural={Automated Guided Vehicles}]{agv}{AGV}{automated guided vehicle}
\newacronym{rfid}{RFID}{Radio-Frequency Identification}

\newacronym[longplural={time-of-flight cameras}]{tofc}{ToF camera}{time-of-flight camera}
\newacronym{pmd}{PMD}{Photonic Mixing Device}
\newacronym{roi}{ROI}{Region of Interest}
\newacronym{iou}{IoU}{Intersection over Union}
\newacronym{ar}{AR}{Augmented Reality}
\newacronym[longplural={Light Detection And Ranging}]{lidar}{LiDAR}{Light Detection And Ranging}
\newacronym[longplural={Frames per Second}]{fps}{FPS}{Frame per Second}
\newacronym{ocr}{OCR}{Optical Character Recognition}
\newacronym{gso}{GSO}{Google Scanned Objects}
\newacronym{lld}{LLD}{Large Logo Dataset}
\newacronym{hog}{HOG}{Histogram of Oriented Gradients}
\newacronym{msssim}{MS-SSIM}{Multiscale Structural Similarity}
\newacronym{cwssim}{CW-SSIM}{Complex Wavelet Structural Similarity}
\newacronym{ssim}{SSIM}{Structural Similarity}
\newacronym{lpips}{LPIPS}{Learned Perceptual Image Patch Similarity}
\newacronym{mae}{MAE}{Mean Absolute Error}
\newacronym{mse}{MSE}{Mean Squared Error}
\newacronym{rocauc}{ROC-AUC}{Area Under the Curve of the Receiver Operating Characteristic}
\newacronym{auc}{AuC}{Area under the Curve}
\newacronym{roc}{ROC}{Receiver Operating Characteristic}
\newacronym{ai}{AI}{Artificial Intelligence}
\newacronym{ml}{ML}{Machine Learning}
\newacronym{dl}{DL}{Deep Learning}
\newacronym{ransac}{RANSAC}{Random Sampling Consensus}
\newacronym[longplural={Artificial Neural Networks}]{nn}{ANN}{Artificial Neural Network}
\newacronym[longplural={Convolutional Neural Networks}]{cnn}{CNN}{Convolutional Neural Network}
\newacronym[longplural={Graph Convolutional Neural Networks}]{gcn}{GCN}{Graph Convolutional Neural Network}
\newacronym[longplural={Graph Neural Networks}]{gnn}{GNN}{Graph Neural Network}
\newacronym[longplural={Support Vector Machines}]{svm}{SVM}{Support Vector Machine}
\newacronym{dbscan}{DBSCAN}{Density-Based Spatial Clustering of Applications with Noise}
\newacronym{sgdm}{SGD+M}{Stochastic Gradient Descent with Momentum}
\newacronym{hmm}{HMM}{Hidden Markov Model}
\newacronym{hci}{HCI}{Human-Computer-Interaction}

\newacronym{eu}{EU}{European Union}

\newglossaryentry{poc}{name={proof of concept}, description={}}
\newglossaryentry{sota}{name={state-of-the-art}, description={}}

\newcommand{\meme}[2]{\textit{(#1, #2)}}

\makeglossaries
\usepackage[capitalize]{cleveref}
\crefname{section}{Sec.}{Secs.}
\Crefname{section}{Section}{Sections}
\Crefname{table}{Table}{Tables}
\crefname{table}{Tab.}{Tabs.}

\AtEveryBibitem{\clearfield{month}}
\AtEveryBibitem{\clearfield{day}}

\def\wacvPaperID{1322} %
\def\confName{WACV}
\def\confYear{2024}

\begin{document}

\title{
    TAMPAR: \\
    Visual Tampering Detection for Parcel Logistics
    in Postal Supply Chains 
}

\author{Alexander Naumann\vspace{1.5mm}\\
FZI and KIT\\
Karlsruhe, Germany\\
{\tt\small anaumann@fzi.de}
\and
Felix Hertlein\vspace{1.5mm}\\
FZI and KIT\\
Karlsruhe, Germany\\
{\tt\small hertlein@fzi.de}
\and
Laura D\"orr\vspace{1.5mm}\\
FZI and KIT\\
Karlsruhe, Germany\\
{\tt\small doerr@fzi.de}
\and
Kai Furmans\vspace{1.5mm}\\
FZI and KIT\\
Karlsruhe, Germany\\
{\tt\small furmans@kit.edu}\\
\\
}
\maketitle
\begin{abstract}
    Due to the steadily rising amount of valuable goods in supply chains, tampering detection for parcels is becoming increasingly important.
    In this work, we focus on the use-case last-mile delivery, where only a single RGB image is taken and compared against a reference from an existing database to detect potential appearance changes that indicate tampering.
    We propose a tampering detection pipeline that utilizes keypoint detection to identify the eight corner points of a parcel.
    This permits applying a perspective transformation to create normalized fronto-parallel views for each visible parcel side surface.
    These viewpoint-invariant parcel side surface representations facilitate the identification of signs of tampering on parcels within the supply chain, since they reduce the problem to parcel side surface matching with pair-wise appearance change detection.
    Experiments with multiple classical and deep learning-based change detection approaches are performed on our newly collected TAMpering detection dataset for PARcels, called \dstamp{}.
    We evaluate keypoint and change detection separately, as well as in a unified system for tampering detection. 
    Our evaluation shows promising results for keypoint (\kpap{} $75.76$) and tampering detection ($81\%$ accuracy, F1-Score $0.83$) on real images.
    Furthermore, a sensitivity analysis for tampering types, lens distortion and viewing angles is presented.
    Code and dataset are available at \urltampar{}.
\end{abstract}

\section{Introduction}

The amount of valuable goods within postal supply chains is increasing steadily \cite{nocetiMulticameraSystemDamage2018}.
This trend implies the rising importance of safety and security considerations in transportation networks.
One crucial aspect to improve safety and security along the supply chain is checking parcels for damages or signs of tampering \cite{naumannParcel3DShapeReconstruction2023}.
Tampering detection, on which we focus in this work, tries to verify and guarantee the integrity of a parcel within the supply chain.
Common cases of potential tampering are applying new tape after opening a parcel or attaching labels to or removing labels from a parcel.
Of course, not all cases where such changes occur are relevant for safety and security considerations, \eg new tape might be applied to prevent items from falling out of a damaged parcel.
\begin{figure}
    \centering
    \includegraphics[width=0.95\linewidth]{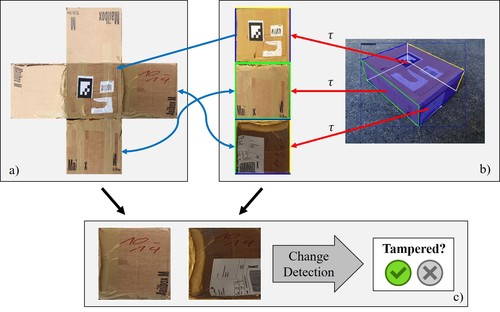}
    \caption{
        We detect tampering by comparing the full parcel texture from a database (a) with the viewpoint-invariant parcel side surfaces of a single image by exploiting parcel corner point predictions (b).
        Appearance change detection is performed for each pair of matching parcel side surfaces to identify tampering (c).
    }
    \label{fig:overview}
\end{figure}

In general, tampering detection for parcels requires a two-step pipeline:
(1) We need to reliably identify a parcel by either its shipping label or unique visual cues on the packaging.
Especially in scenarios with numerous visually similar parcels, using the latter can be challenging, while at the same time, the shipping label might not always be visible.
(2) We need to compare the appearance of the packaging across time to detect changes that might stem from tampering.
This is a challenging task since the photos used for the comparison can show the objects of interest from different viewing angles and under varying lighting conditions.

In this work, we focus on step (2), since parcel re-identification has been studied by \textcite{clausenParcelTrackingDetection2019}, and present an approach for tampering detection for already identified parcels (\cf \cref{fig:overview}).
Since last-mile delivery is considered as use-case, we assume that only a single RGB image is available, which should be compared against a reference from an existing database.
Similar to \textcite{nocetiMulticameraSystemDamage2018}, appearance change detection is performed separately per parcel side surface, and if at least one of the parcel side surfaces has been tampered with, the whole parcel is considered to have undergone tampering.
To tackle this problem, we suggest the usage of keypoint detection to identify the parcel corners as a first step toward change detection.
Knowledge of the eight corner points of cuboid-shaped parcels enables computing normalized fronto-parallel views of all visible parcel side surfaces by applying a perspective transformation $\tau$ (\cf \cref{fig:overview} (b)).
These views eliminate the viewpoint variance and thus, facilitate change detection and potentially also re-identification of parcel side surfaces.
We use the \dsparcel{} dataset \cite{naumannParcel3DShapeReconstruction2023} to demonstrate the capabilities of keypoint detectors for generating viewpoint-invariant parcel side surface representations from single RGB images.
Additionally, we collected a novel dataset for change detection in postal supply chains, and present a detailed analysis of change detection approaches for tampering detection.
The main contributions of our work are as follows
\begin{itemize}[itemsep=0pt]
    \item we suggest an effective keypoint ordering for parcel detection and side surface segmentation,
    \item we present \dstamp{}, a novel dataset for TAMpering detection of PARcels, and
    \item we propose and evaluate an approach for tampering detection, which exploits keypoint and change detection. %
\end{itemize}

\section{Related Work}
\label{sec:related_work}
We review related literature in logistics applications, 3D bounding box detection, keypoint estimation and change detection in the following.

\paragraph*{Applications in Logistics.}
\textcite{karacaMulticameraVisionSystem2005} present an early approach using a stereo camera and feature matching techniques to track parcels along a conveyor belt.
\textcite{clausenParcelTrackingDetection2019} present an approach for tracking parcels inside a logistics facility.
A \maskrcnn{}-based \cite{heMaskRCNN2017} Siamese network \cite{ bromleySignatureVerificationUsing1993} complemented with their so-called feature improver head is used to re-identify parcels.
They create a manually labeled dataset of 3,306 images taken by 37 different cameras with a total of 14,248 parcels.
The evaluation shows that currently around \SI{81}{\%} of parcels are tracked correctly.
For more details on literature regarding re-identification we refer to \textcite{yeDeepLearningPerson2022} and \textcite{khanSurveyAdvancesVisionbased2019}.
\textcite{naumannRefinedPlaneSegmentation2020} work towards parcel side surface segmentation.
By combining plane segmentation \cite{liuPlaneRCNN3DPlane2019} and contour detection \cite{cannyComputationalApproachEdge1986,soriaDenseExtremeInception2020}, they present an approach to refine parcel side surface segmentation masks without relying on any task-specific training data.
\textcite{naumannParcel3DShapeReconstruction2023} tackle the problem of estimating the 3D shape of potentially damaged parcels from a single RGB input.
They extend \cubercnn{} \cite{brazilOmni3DLargeBenchmark2023} by an iterative mesh refinement \cite{gkioxariMeshRCNN2019} and present \dsparcel{}, a dataset comprising over 13,000 images of cuboid-shaped and damaged parcels with full 3D annotations.
\textcite{nocetiMulticameraSystemDamage2018} present a multi-camera system for damage and tampering detection in postal supply chains.
Damages are detected by finding the parallelepiped which best aligns with the captured images.
For tampering detection a \gls*{hog} \cite{dalalHistogramsOrientedGradients2005} for the parcel side surfaces is used.
Rotation invariance is accomplished by considering all possible rotations with histogram intersection as similarity measure.
Tampering is reported when the similarity of two feature vectors is below a certain threshold. 
Other works focusing on parcels consider synthetic training data generation \cite{naumannScrapeCutPasteLearn2022}, tracking inside a moving truck \cite{huCuboidDetectionTracking2021} and depalletization \cite{arpentiRGBDRecognitionLocalization2020,chiaravalliIntegrationMultiCameraVision2020}. Finally, \textcite{naumannLiteratureReviewComputer2023} present a detailed overview of computer vision applications in transportation logistics and warehousing.

\paragraph*{3D Bounding Box Detection.}

\textcite{dwibediDeepCuboidDetection2016} present an early deep learning-based approach to estimate the 3D bounding box of cuboid-shaped objects.
Generally, 3D bounding box detection is a common task for autonomous driving \cite{arnoldSurvey3DObject2019}.
Approaches often rely on only estimating yaw, since they can exploit the fact that vehicles are driving on the road.
\textcite{liRTM3DRealTimeMonocular2020} exploit 2D/3D correspondences by estimating keypoints of cars to improve 3D bounding box detection.
\textcite{ruiGeometryConstrainedCarRecognition2020} introduce a framework for vehicle recognition from a single RGB image.
They estimate a 3D bounding box which is used to compute normalized views for the front, side and roof view of a car by applying a perspective transformation.
This information is fused with region-aligned features of the respective region of interest to estimate the vehicle model.

\paragraph*{Keypoint Detection.}

Lots of research tackling keypoint estimation considers monocular human pose estimation, which is reviewed by \textcite{chenMonocularHumanPose2020} and \textcite{chen2DHumanPose2022}.
\textcite{dorrTetraPackNetFourCornerBasedObject2021} treat the problem of packaging structure recognition.
The goal is to identify the number, type and arrangement of small load carriers on a uniformly packed transport unit from a single RGB image.
They extend CornerNet \cite{lawCornerNetDetectingObjects2018} to leverage keypoint estimation to detect objects based on four arbitrary corner points. 

\paragraph*{Change Detection.}

To detect signs of tampering, after re-identification, change detection is necessary.
Change detection is most commonly applied for remote-sensing and street views and reviewed by \textcite{shiChangeDetectionBased2020}.
A dataset for change detection in industrial environments has been presented by \textcite{parkChangeSimEndtoEndOnline2021}.
Furthermore, \textcite{parkDualTaskLearning2022} propose the novel change detection approach SimSaC which is targeted towards industrial use-cases.
SimSaC relies on dual task learning and exploits both, dense correspondence and mis-correspondence to increase robustness when encountering imperfect matches.

~\\
While \textcite{nocetiMulticameraSystemDamage2018} also tackle the problem of tampering detection, they focus on a constrained environment with calibrated background, constant illumination and a multisensory setup.
In contrast to that, our approach does not have any such constraints and relies just on a single RGB image as input.
Consequently, ours is the only approach suitable for scenarios such as last-mile delivery and cannot be fairly compared to the work by \textcite{nocetiMulticameraSystemDamage2018}.
Furthermore, we rely on existing keypoint and change detection approaches and strive to combine them efficiently, however, we do not aim to develop novel approaches in these areas.

\section{Approach}
\label{sec:approach}

We present our approach for parcel keypoint detection in \cref{sec:approach:keypoint} and for change detection in \cref{sec:approach:change}.
Details on our novel dataset \dstamp{} are given in \cref{sec:approach:dataset}.

\subsection{Parcel Keypoint Detection}
\label{sec:approach:keypoint}

We use a \maskrcnn{} \cite{heMaskRCNN2017} with keypoint head and a \resnetfpn{} \cite{heDeepResidualLearning2016,linFeaturePyramidNetworks2017} backbone for our experiments.
This choice is motivated by the fact, that we do not focus on improving keypoint detection techniques, but rather want to demonstrate the usefulness of well-established baselines for the use-case of parcel corner detection.

\begin{figure}[h!]
    \centering
    \includegraphics[width=0.38\linewidth]{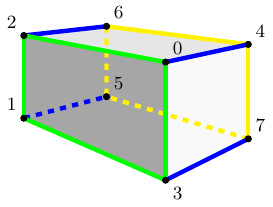}
    ~~
    \includegraphics[width=0.38\linewidth]{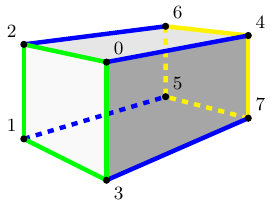}
    \includegraphics[width=0.38\linewidth]{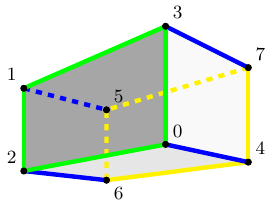}
    ~~
    \includegraphics[width=0.38\linewidth]{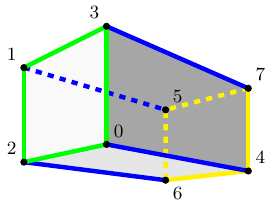}
    \caption{
        Visualization of the consistent and unambiguous keypoint ordering for a cuboid without well-defined front and back across different viewing angles.
        We highlight the front side in green and the back side in yellow.%
    }
    \label{fig:approach:keypoints}
\end{figure}

One key challenge for this use-case is to identify an unambiguous keypoint ordering which works well with \acrlongpl*{nn} since there are several options for ordering keypoints of a parcel.
In contrast to the common application of 3D bounding box detection for autonomous driving, where vehicles have a well-defined front and back side, there is no such notion for parcels.
To have a consistent, unambiguous keypoint ordering with explicit visual cues, we proceed as follows.
We assume, that three parcel side surfaces are visible in each image and define the front of a parcel by choosing the visible parcel side surface whose normal aligns best with a left- and front-facing vector, \idest $(x, y, z) = (1, 0, -0.5)$.
From this, we derive our keypoint ordering definition, which is visualized in \cref{fig:approach:keypoints} and described in the following.
We denote the number of visible $\alpha$ and invisible $\beta$ parcel side surfaces that intersect in keypoint \kpi{i} as \kpi{i}=\vuss{$\alpha$}{$\beta$}. %
On the front side (highlighted in green in \cref{fig:approach:keypoints}), we define the keypoints:
\begin{itemize}[itemsep=-2.5pt]
    \setcounter{enumi}{-1}
    \item \kpi{0} = \vuss{3}{0}: point of intersection of the three visible parcel side surfaces, which is located inside the convex hull of the parcel.
    \item \kpi{1}=\vuss{1}{2}: joint point of the two invisible parcel side surfaces, where only two visible parcel edges intersect.
    \item \kpi{2}=\vuss{2}{1}, leftmost: leftmost point of the remaining points, where three visible parcel edges intersect. %
    \item \kpi{3}=\vuss{2}{1}, rightmost: remaining point, which is the rightmost point that belongs to two visible parcel side surfaces and one invisible one.
\end{itemize}

The backside (highlighted in yellow in \cref{fig:approach:keypoints}) of the parcel is the one across from the front side, and we define the keypoint order as follows:
\begin{itemize}[itemsep=-2.5pt]
    \setcounter{enumi}{3}
    \item \kpi{4}=\vuss{2}{1}: point that is part of two visible parcel side surfaces and thus, at this point three visible parcel edges intersect.
    \item \kpi{5}=\vuss{0}{3}: self-occluded keypoint, which is the point of intersection of the three invisible parcel side surfaces.
    \item \kpi{6}=\vuss{1}{2}, leftmost: leftmost point of the remaining points, where two visible parcel edges intersect.
    \item \kpi{7}=\vuss{1}{2}, rightmost: remaining point, which is the rightmost point where two visible edges intersect.
\end{itemize}

This keypoint ordering is used for training and evaluating corner point detection in the following.
Note, that it is not invariant to horizontal, but only to vertical flipping of the image. %
Furthermore, technically, estimating seven keypoints would be sufficient to infer all eight, however, we want to show that the estimation even works for the self-occluded keypoint \kpi{5}.
The information on the seven visible keypoints can be utilized to compute viewpoint-invariant parcel side surface representations by applying a perspective transformation.
This, in turn, enables the composition of parcel texture mappings as visualized in \cref{fig:overview} (a).

\subsection{Change Detection}
\label{sec:approach:change}

In our use-case, we assume that a postman takes a single image of a parcel which seems suspicious of potential tampering.
First, the parcel keypoints are extracted and the viewpoint-invariant parcel side surfaces of size $400 \times 400$ pixels are computed as described in \cref{sec:approach:keypoint} and visualized in \cref{fig:overview} (b). %
By exploiting this information, we can reduce the task of tampering detection of parcels to comparing fronto-parrallel parcel side surface representations.
If one parcel side surface has been tampered with, the parcel is considered tampered.

While the usage of viewpoint-invariant representations alleviates the problem of perspective distortion, change detection remains challenging since image alignment issues cannot fully be resolved, and additionally, the lighting might vary significantly (\cf \cref{fig:overview} (c)).
To cope with these issues, we use SimSaC \cite{parkDualTaskLearning2022}.
SimSaC is a recent approach for robust change detection with imperfect matches.
It estimates scene flow using correspondence maps at the same time as change masks by exploiting mis-correspondences.
This enables robustness against geometric transformations and differences in lighting.

\begin{figure*}[h!]
    \centering
    \begin{subfigure}[b]{0.14\linewidth}
        \centering
        \includegraphics[width=\linewidth]{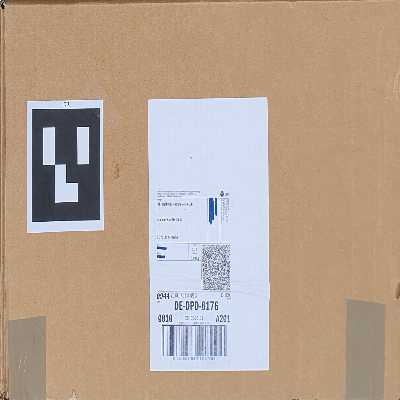}
        \includegraphics[width=\linewidth]{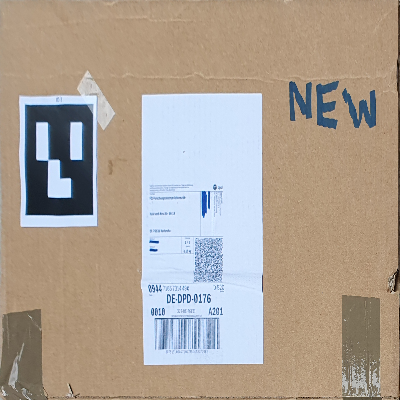}
        \caption{None}
        \label{fig:methods:a}
    \end{subfigure}
    \begin{subfigure}[b]{0.14\linewidth}
        \centering
        \includegraphics[width=\linewidth]{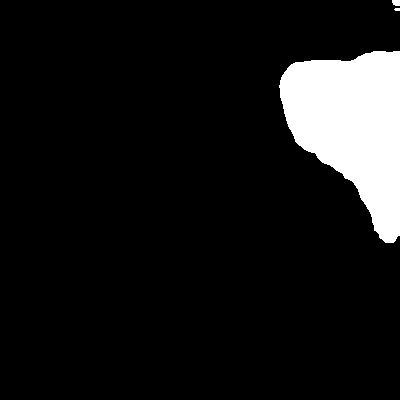}
        \includegraphics[width=\linewidth]{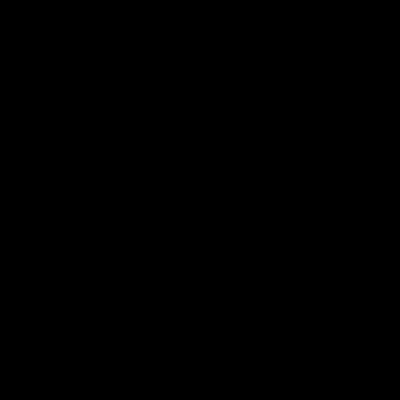}
        \caption{SimSaC}
        \label{fig:methods:b}
    \end{subfigure}
    \begin{subfigure}[b]{0.14\linewidth}
        \centering
        \includegraphics[width=\linewidth]{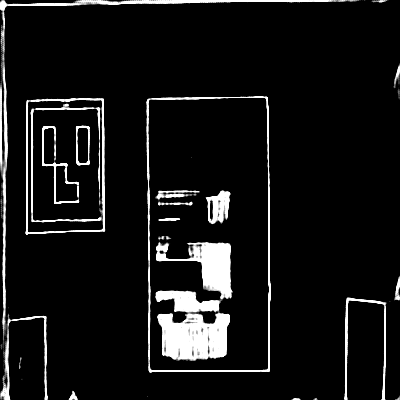}
        \includegraphics[width=\linewidth]{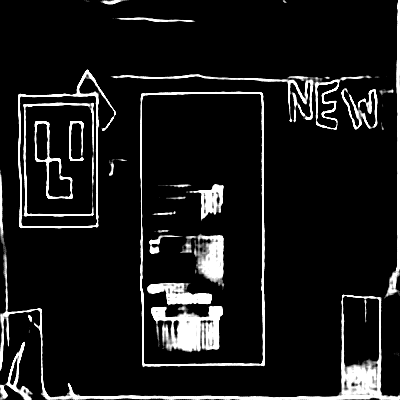}
        \caption{DexiNed}
        \label{fig:methods:c}
    \end{subfigure}
    \begin{subfigure}[b]{0.14\linewidth}
        \centering
        \includegraphics[width=\linewidth]{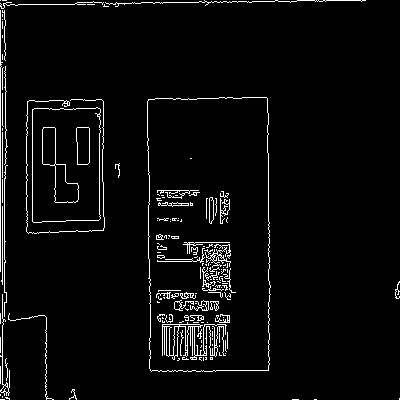}
        \includegraphics[width=\linewidth]{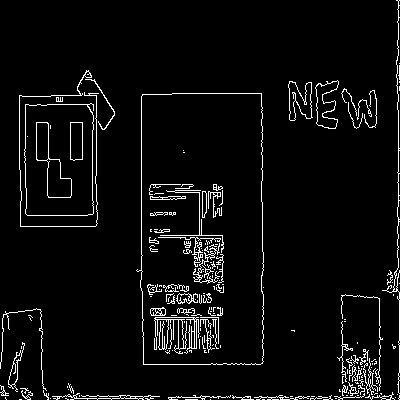}
        \caption{Canny}
        \label{fig:methods:d}
    \end{subfigure}
    \begin{subfigure}[b]{0.14\linewidth}
        \centering
        \includegraphics[width=\linewidth]{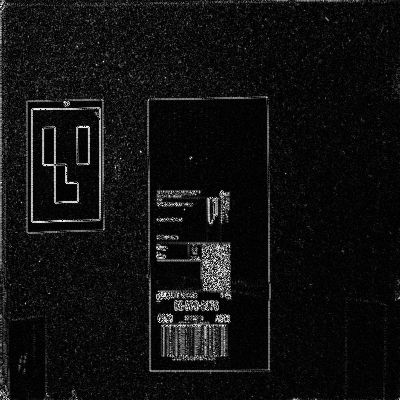}
        \includegraphics[width=\linewidth]{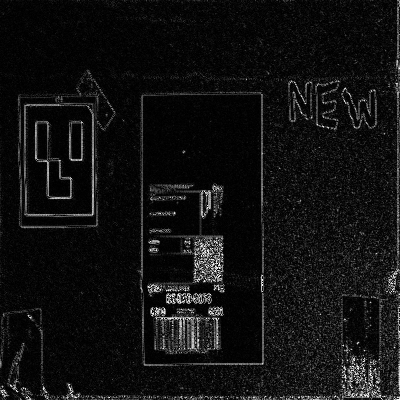}
        \caption{Laplac.}
        \label{fig:methods:e}
    \end{subfigure}
    \begin{subfigure}[b]{0.14\linewidth}
        \centering
        \includegraphics[width=\linewidth]{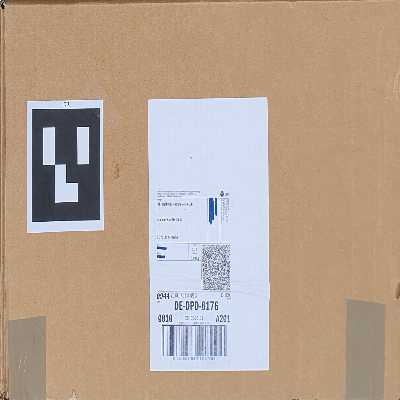}
        \includegraphics[width=\linewidth]{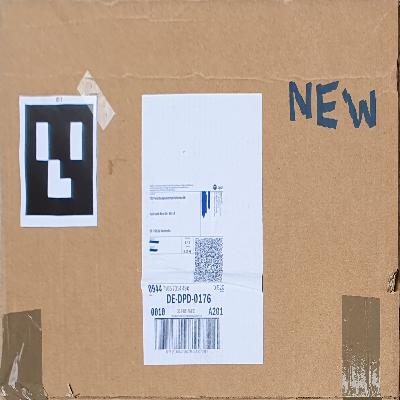}
        \caption{Mean Ch.}
        \label{fig:methods:f}
    \end{subfigure}
    \caption{
        Examples of the different image homogenization methods before (top) and after (bottom) tampering.
        Note that SimSaC \cite{parkDualTaskLearning2022} is the only approach that localizes potential tampering and directly outputs change maps.
    }
    \label{fig:methods}
\end{figure*}

We benchmark SimSaC against several baselines, each combining an image homogenization approach and a similarity metric.
For image homogenization, we utilize (\cf \cref{fig:methods})
\begin{itemize}[itemsep=-2.5pt]
    \item \textit{DexiNed}: Dense EXtreme Inception Network for Edge Detection \cite{soriaDenseExtremeInception2020}
    \item \textit{Canny}: Adaptive Canny edge detection \cite{cannyComputationalApproachEdge1986,jieImprovedAdaptiveThreshold2012}
    \item \textit{Laplacian}: Laplacian filter
    \item \textit{Mean Channel}: Per-channel mean alignment
\end{itemize}
As image similarity metrics, we consider
\begin{itemize}[itemsep=-2.5pt]
    \item \gls*{lpips} \cite{zhangUnreasonableEffectivenessDeep2018},
    \item \gls*{ssim} \cite{wangImageQualityAssessment2004},
    \item \gls*{msssim} \cite{wangMultiscaleStructuralSimilarity2003},
    \item \gls*{cwssim} \cite{sampatComplexWaveletStructural2009},
    \item \gls*{hog} \cite{dalalHistogramsOrientedGradients2005} feature similarity \footnote{We use 9 orientation bins, 8 pixels per cell and 2 cells per block.}, and
    \item \gls*{mae}.
\end{itemize}

A change is detected when the input and reference parcel side surface image after applying the image homogenization to both, have a low image similarity.
Suitable thresholds for image similarity will be determined in \cref{sec:eval:tampering}.

\subsection{Dataset}
\label{sec:approach:dataset}

Our dataset resembles a use-case with multisensory setups within logistics facilities and a simple cell phone camera during the last-mile delivery.
More precisely, we assume that multiple cameras are used to capture and segment all five visible parcel side surfaces in logistics facilities. 
Note that we also suppose that the side surface with the unique identifier is always visible, which means that the opposing side surface is never visible.
Subsequently, the parcel ID and texture map, as visualized in \cref{fig:overview} (a), are saved to a database.
Finally, a single RGB image of a parcel with suspected tampering is taken during last-mile delivery and compared against its high-quality reference texture.

To generate a suitable dataset for this use-case, we proceed as follows.
We use ArUco markers to uniquely identify parcels and the spatial relationships between their side surfaces.
The parcel textures for the database are generated by taking several images of the parcel in its original state, \idest without tampering.
By manually labeling the parcel corner points, we automatically generate the full parcel texture by applying perspective transformations.
Subsequently, we apply different types of tampering to three out of the five relevant parcel side surfaces.
While real-world tampering attempts focus on a single parcel side surface, our dataset design enables a more diverse analysis of tampering detection by considering a larger number of tampering examples.
As mentioned before, transferring side surface tampering to the parcel level is straightforward.
We consider three different types of tampering flags, each with an \textit{easy} and a \textit{hard} to detect variant:

\begin{itemize}[itemsep=-2.5pt]
    \item \textit{Label}: Adding a new shipping label (\textit{easy}) or transportation hints (\textit{hard})
    \item \textit{Tape}: Adding new tape, which covers more than 50\% of the longer side (\textit{easy}), or less than 25\% of the shorter side (\textit{hard})
    \item \textit{Writing}: Adding manually written text, using a pen with \SI{5}{}-\SI{15}{mm} (\textit{easy}) or \SI{1.5}-\SI{3}{mm} of width (\textit{hard})
\end{itemize}

Note that adding written text usually would not be considered tampering.
However, we strive to detect diverse appearance changes to reliably flag parcels for manual inspection.
In total, we collect and annotate $296$ images of $10$ parcels for the training/validation and $614$ images of $20$ parcels for the test set.
Since each image contains three visible parcel side surfaces, \dstamp{} comprises $888$ images for training/validation and $1842$ images for testing change detection.
The main difference to existing datasets such as \dsparcel{} \cite{naumannParcel3DShapeReconstruction2023} and \dsreal{} \cite{naumannScrapeCutPasteLearn2022} is that we have paired images of the same parcel across different points in time, \idest before and after tampering.

\section{Evaluation}
\label{sec:eval}

We first evaluate keypoint detection for parcel corners separately in \cref{sec:eval:kp}.
Subsequently, we evaluate the considered change detection approaches isolated and in combination with keypoint estimation in \cref{sec:eval:tampering}.
Furthermore, we present a sensitivity analysis on the influence of the tampering type, lens distortion and viewing angles.

\subsection{Parcel Corner Point Estimation}
\label{sec:eval:kp}

For all experiments, we use a \resnetfpn{} \cite{heDeepResidualLearning2016,linFeaturePyramidNetworks2017} that was pre-trained on MS COCO \cite{linMicrosoftCOCOCommon2014} as backbone and freeze its weights at stage four.
We use \gls*{sgdm} with a batch size of 16 and a cosine learning rate schedule \cite{loshchilovSGDRStochasticGradient2017}.
The initial learning rate is set to \SI{0.001}{} and the final learning rate to \SI{0}{} after \SI{10000}{} iterations.
Moreover, a linear warm-up during the first \SI{1000}{} iterations is applied.

Training is always performed on the synthetic dataset \dsparcel{} \cite{naumannParcel3DShapeReconstruction2023} which contains cuboid-shaped and damaged parcel images.
For the evaluation, we consider synthetic and real-world data in the following.
We evaluate bounding box detection, instance segmentation and keypoint detection, and summarize the quantitative results in \cref{table:eval:aps}.

For the evaluation of keypoint detection using \kpap{}\footnote{See \href{https://cocodataset.org/\#keypoints-eval}{https://cocodataset.org/\#keypoints-eval} for details.}, it is necessary to define $\kappa_i$ for each keypoint.
This value is usually obtained by comparing redundantly annotated images to infer each keypoints' annotation precision.
Since no redundantly annotated images are available, we select $\kappa_5=0.1$ for the self-occluded and $\kappa_i = 0.05,~ i \in \{0,1,2,3,4,6,7\}$ for the visible keypoints, 
which is close to the $\kappa$ for human hips ($0.107$) and human wrists ($0.062$), respectively \cite{linMicrosoftCOCOCommon2014}.
We argue that human wrists are a suitable approximation because the keypoints for \dsparcel{} and \dsreal{} are computed from 3D bounding boxes, frequently leading to a misalignment between the parcel corners in the image and the actual annotated keypoints.
This misalignment is also present for damaged parcels, where the keypoints correspond to the ones of the pristine version of the parcel.

\begin{table*}[h!]
    \centering
    \begin{tabular}{lccccccccc}
        \toprule
                          & \multicolumn{2}{c}{Box} & \multicolumn{2}{c}{Mask} & \multicolumn{2}{c}{Keypoint}                                           \\ 
        Dataset           & \ap{}                   & \ap{75}                  & \ap{}                        & \ap{75}     & \ap{}       & \ap{75}     \\
        \midrule
        \dsparcel{}       & 93.62 (0.1)             & 98.46 (0.2)              & 97.54 (0.2)                  & 98.58 (0.3) & 88.80 (0.2) & 94.06 (0.2) \\
        \dsreal{}~        & 84.88 (0.2)             & 97.28 (0.1)              & 85.02 (0.2)                  & 96.92 (0.6) & 75.76 (0.5) & 85.36 (1.2) \\
        \dstamp{} (ours)~ & 96.38 (0.2)             & 99.72 (0.5)              & 98.94 (0.2)                  & 99.70 (0.5) & 97.18 (0.5) & 99.12 (0.4) \\
        \bottomrule
    \end{tabular}
    \caption{
        Quantitative performance analysis of the \resnetfpn{} for bounding box detection, instance segmentation and keypoint detection.
        We repeated all trainings five times and report \textit{mean (standard deviation)}.
    }
    \label{table:eval:aps}
\end{table*}

\subsubsection{Synthetic Data}
\label{sec:eval:kp:synthetic}

The quantitative results from \cref{table:eval:aps} indicate excellent performance for bounding box detection and instance segmentation, with a \boxap{} of $93.62$ and a \maskap{} of $97.54$.
Likewise, keypoint detection achieves strong results with a \kpap{} of $88.80$.

\begin{figure}[h]
    \centering
    \begin{subfigure}[b]{0.45 \linewidth}
        \centering
        \includegraphics[width=.47\linewidth]{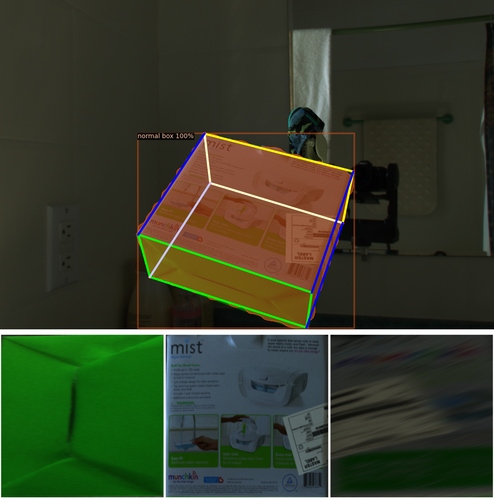}
        \includegraphics[width=.47\linewidth]{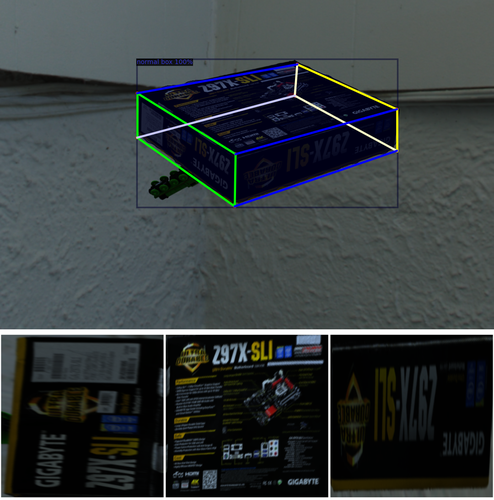}
        \caption{Intact Parcels}
        \label{fig:eval:synthetic:a}
    \end{subfigure}%
    ~%
    \begin{subfigure}[b]{0.45 \linewidth}
        \centering
        \includegraphics[width=.47\linewidth]{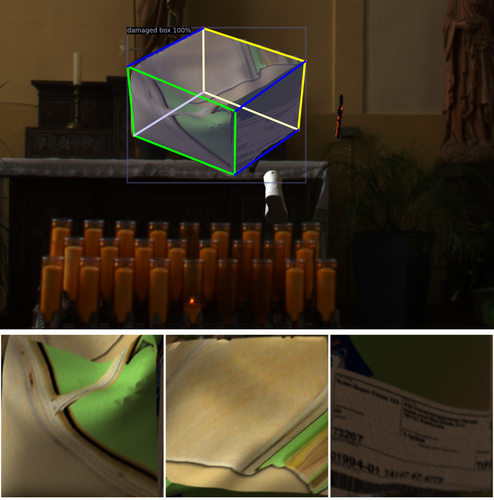}
        \includegraphics[width=.47\linewidth]{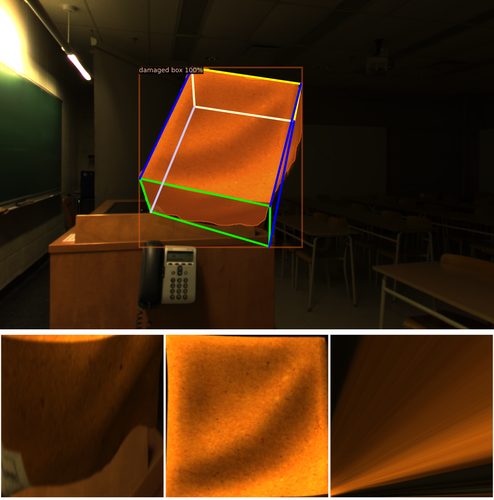}
        \caption{Damaged Parcels}
        \label{fig:eval:synthetic:b}
    \end{subfigure}
    \begin{subfigure}[b]{0.75\linewidth}
        \centering
        \includegraphics[width=.31\linewidth]{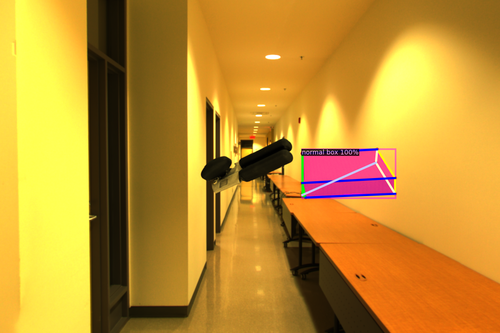}
        ~
        \includegraphics[width=.31\linewidth]{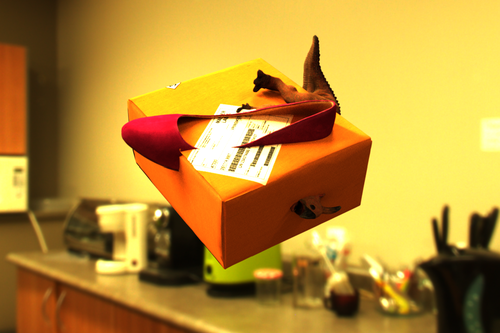}
        \caption{Problematic Cases}
        \label{fig:eval:synthetic:c}
    \end{subfigure}
    \caption{
        Exemplary qualitative results for synthetic parcels. 
    }
    \label{fig:eval:synthetic}
\end{figure}

Qualitative examples are presented in \cref{fig:eval:synthetic}.
For intact, \idest cuboid-shaped, parcels keypoint detection enables computing high-quality fronto-parallel views of the parcel side surfaces as can be seen in \cref{fig:eval:synthetic:a}.
Strong distortions of parcel side surface views, however, cannot be recovered and lead to low-quality representations, which are challenging to use for tampering detection.
In the case of damaged parcels, the representations' quality strongly depends on the degree of deformation (\cf \cref{fig:eval:synthetic:b}).
Strong deformations also impede tampering detection.
Problematic cases can include imprecise or missing detections (\cf \cref{fig:eval:synthetic:c}). %

\subsubsection{Real Data}
\label{sec:eval:kp:real}

Due to the fact that training was only performed on the synthetic training dataset \dsparcel{} \cite{naumannParcel3DShapeReconstruction2023}, a domain gap occurs when evaluating on the two real-world datasets \dsreal{} \cite{naumannScrapeCutPasteLearn2022} and \dstamp{}.
This domain gap manifests itself in the generally lower performance on \dsreal{} compared to the evaluation on synthetic data, as seen in \cref{table:eval:aps}.
At the same time, performance on \dstamp{} is higher, presumably due to the simpler nature of the dataset - all images are high-quality and show only a single parcel in the center.
Performance for bounding box detection and instance segmentation remains high with a \boxap{} of $84.88$/$96.38$ and a \maskap{} of $85.02$/$98.94$, on \dsreal{} and \dstamp{}, respectively.
The same holds true for the performance of keypoint detection, which reaches $75.76$ and $97.18$ \kpap{}.

\begin{figure}[h!]
    \centering
    \begin{subfigure}[b]{0.87\linewidth}
        \centering
        \includegraphics[width=.47\linewidth]{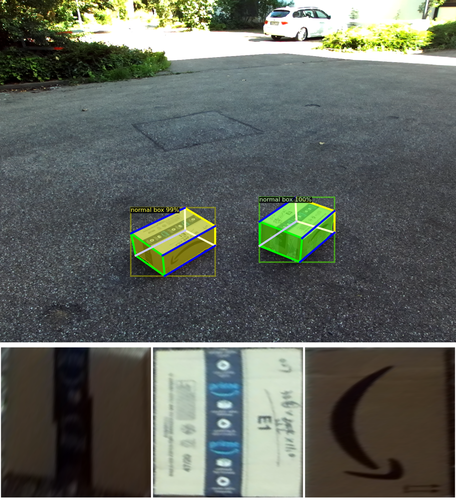}
        \includegraphics[width=.47\linewidth]{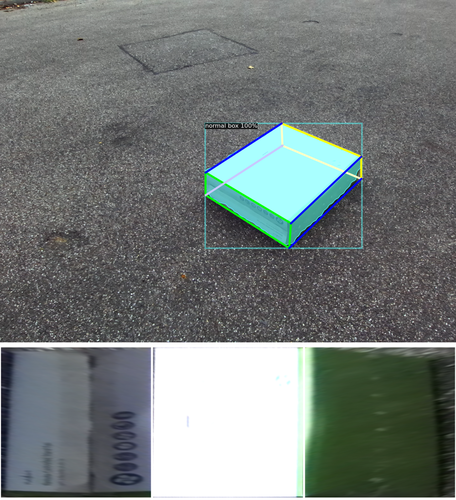}
        \caption{Intact Parcels}
        \label{fig:eval:real:a}
    \end{subfigure}
    \begin{subfigure}[b]{0.87\linewidth}
        \centering
        \includegraphics[width=.47\linewidth]{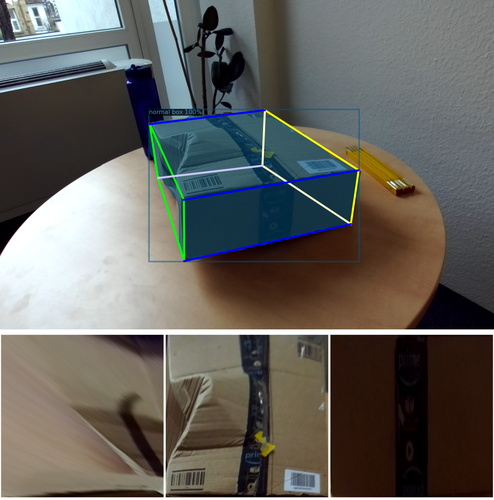}
        \includegraphics[width=.47\linewidth]{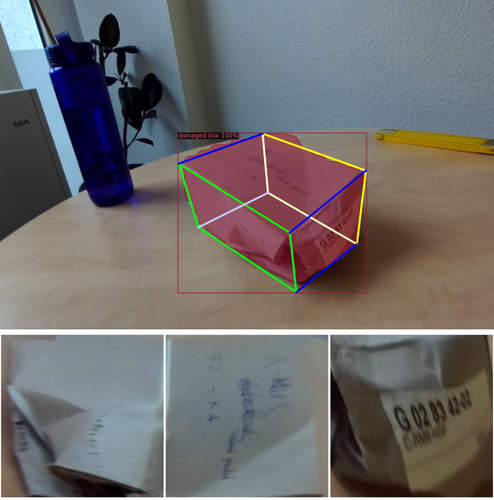}
        \caption{Damaged Parcels}
        \label{fig:eval:real:b}
    \end{subfigure}
    \begin{subfigure}[b]{\linewidth}
        \centering
        \includegraphics[width=0.3\linewidth]{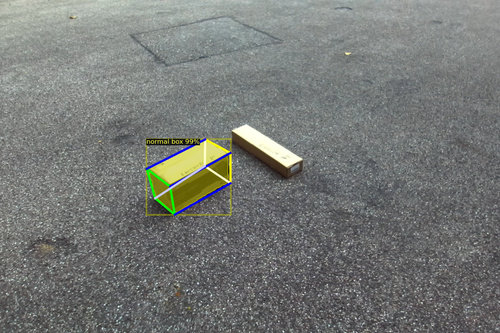}
        ~
        \includegraphics[width=.3\linewidth]{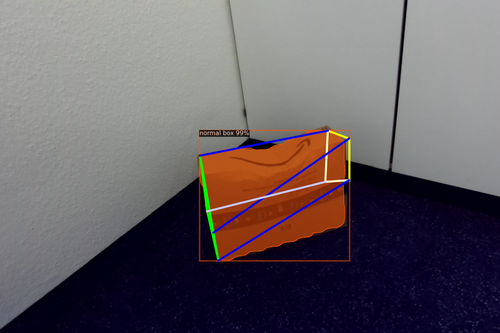}
        ~
        \includegraphics[width=0.3\linewidth]{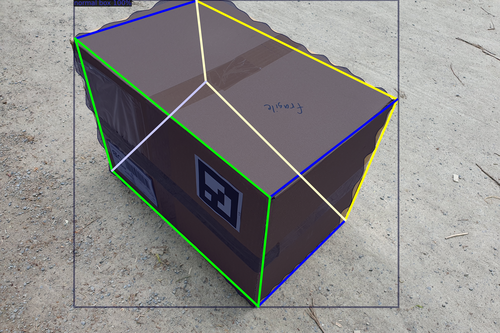}
        \caption{Problematic Cases}
        \label{fig:eval:real:c}
    \end{subfigure}
    \caption{
        Exemplary qualitative results for real parcels. 
    }
    \label{fig:eval:real}
\end{figure}

Quantitative inspection of the prediction results confirms the suitability of \dsparcel{} and our proposed keypoint ordering.
Especially for cuboid-shaped parcels, as visualized in \cref{fig:eval:real:a}, results look very promising for applications in tampering detection.
Furthermore, we evaluate our approach on images of damaged parcels without ground truth annotations (\cf \cref{fig:eval:real:b}).
These qualitative impressions also underpin the suitability of our approach, however, detecting keypoints accurately for damaged parcels seems to be a more difficult task.
Examples of failed detections include missing and imprecise keypoint localizations, as visualized in \cref{fig:eval:real:c}.

\subsubsection{Sensitivity Analysis: Lens Distortion}
\label{sec:eval:kp:lens}

\begin{figure}
    \centering
    \includegraphics[width=.25\linewidth]{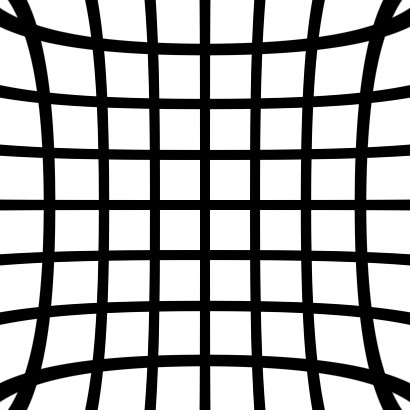}~
    \includegraphics[width=.25\linewidth]{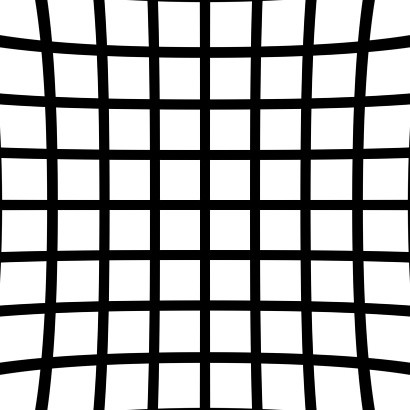}~
    \includegraphics[width=.25\linewidth]{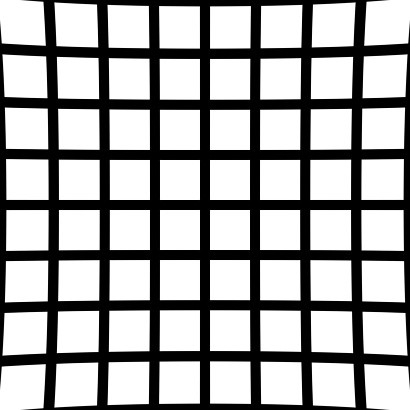}\\
    \includegraphics[width=.25\linewidth]{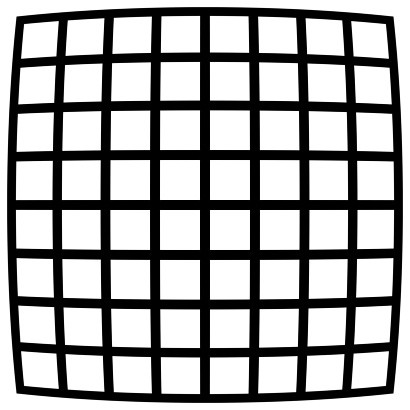}~
    \includegraphics[width=.25\linewidth]{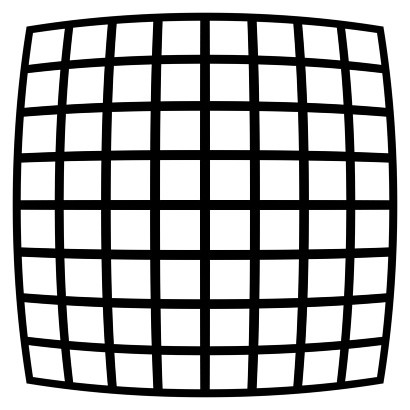}~
    \includegraphics[width=.25\linewidth]{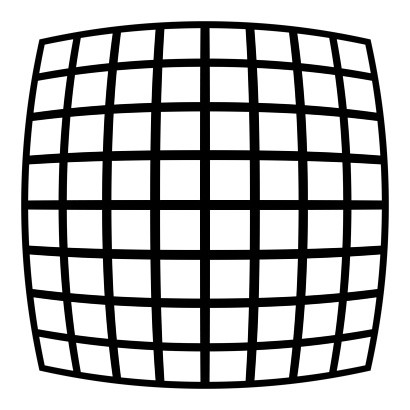}%
    \caption{
        Visualization of the investigated distortion effects with parameter $A \in [-0.08, -0.04, -0.02, 0.04, 0.08, 0.16]$, $B = 0$, $C=0$, and $D=1.0$.
    }
    \label{fig:eval:distortion:samples}
\end{figure}

\begin{figure}
    \centering
    \includegraphics[width=0.9\linewidth]{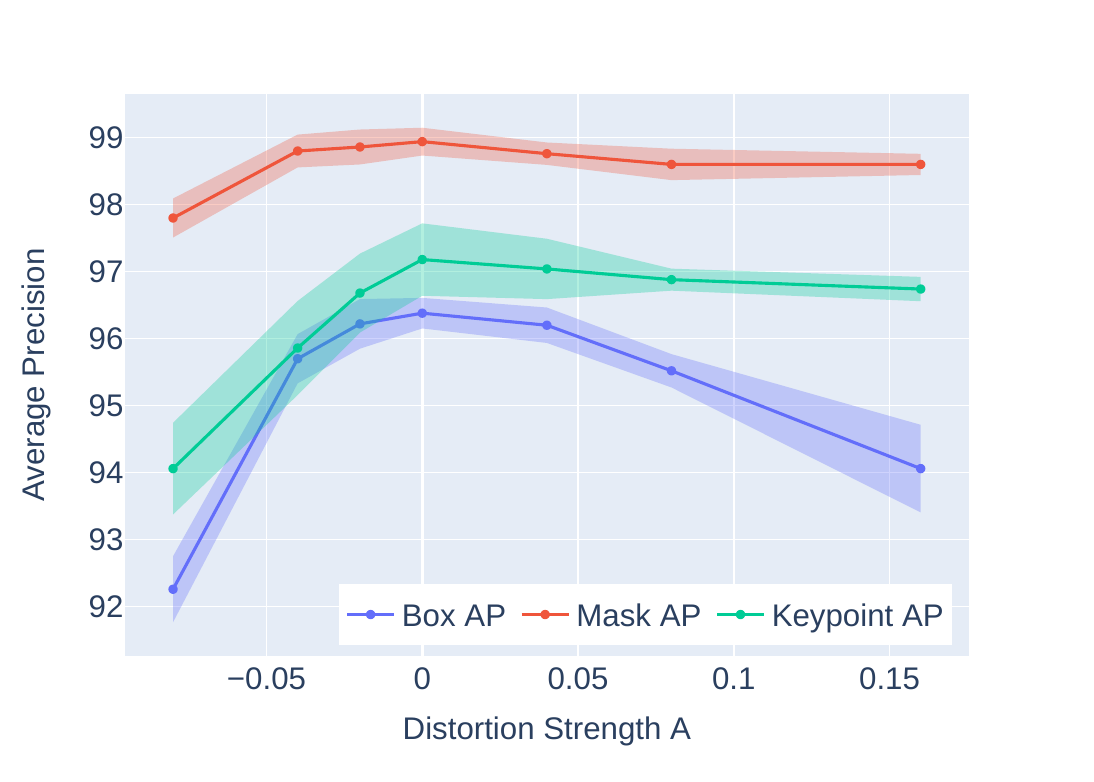}
    \caption{
        Quantitative performance analysis of the \resnetfpn{} under different types of lens distortion.
        We repeated all trainings five times and report mean values with standard deviations as error boundaries.
    }
    \label{fig:eval:distortion:res}
\end{figure}

We investigate the influence of barrel distortion according to
\[
    r_{\text{src}} = r_{\text{dist}} \cdot \left(A \cdot r_{\text{dist}}^3 + B \cdot r_{\text{dist}}^2 + C \cdot r_{\text{dist}} + D \right)
\]
with $r_{\text{src}}$ being the radial distance from the image center in the input image, and $r_{\text{dist}}$ the one, in the distorted output. 
We analyze six different settings, which are visualized in \cref{fig:eval:distortion:samples} by creating distorted dataset versions with parameter $A \in [-0.08, -0.04, -0.02, 0.04, 0.08, 0.16]$, $B = 0$, $C=0$, and $D=1.0$.
Note that these datasets can be smaller in size, since we discard instances if the distortion corrupted the annotations (\eg keypoints lie outside the image) or the ArUco detection.
Results in \cref{fig:eval:distortion:res} indicate that instance segmentation performance is robust \wrt distortion effects.
While keypoint detection only degrades for pincushion distortions ($A<0$), bounding box detection results are also affected by strong barrel distortions ($A>0$).


\subsection{Tampering Detection}
\label{sec:eval:tampering}

We evaluate change detection in isolation by using the ground truth keypoint annotations, as well as in a combined system on our novel dataset \dstamp{} by using keypoint predictions from \cref{sec:eval:kp}.
Furthermore, we analyze the influences of tampering types, lens distortion, and viewing angles.

\subsubsection{Pipeline Evaluation}
\label{sec:eval:tampering:main}

Considering the input and reference parcel we first perform marker-based side surface matching.
Subsequently, we apply the image homogenization to both (input and reference side surface view) and compute their image similarity using all metrics mentioned in \cref{sec:approach:change}.
We denote the combination of image homogenization \textit{Method A} and similarity \textit{Metric B} as \meme{Method A}{Metric B},
and seek to determine the best image similarity metrics and corresponding thresholds. %
SimSaC \cite{parkDualTaskLearning2022} poses a special case since it uses the input and reference image to output change maps.
This enables localization of tampering, which is advantageous in practice, however, not evaluated in this work. 
Instead, we compare the binary change map to a black image (\idest the change map corresponding to no changes) to compute image similarity.
We summarize the evaluation results using simple thresholding by training a decision tree of depth one per method using all similarity metrics as input in \cref{table:eval:tamp:simple}.

\begin{table*}[h!]
    \centering
    \begin{tabular}{llccccc}
        \toprule
        Method    & Metric~~                   & ~Accuracy~                  & ~Precision~                 & ~Recall~           & ~F1-Score~                  & ROC-AUC                     \\
        \midrule
        None      & \gls*{lpips}/\gls*{msssim} & 0.66/0.65                   & 0.66/0.65                   & 0.91/0.93          & 0.76/0.76                   & 0.60/0.58                   \\
        SimSaC    & \gls*{lpips}/\gls*{mae}    & \textbf{0.81}/\textbf{0.80} & \textbf{0.91}/\textbf{0.93} & 0.76/0.72          & \textbf{0.83}/\textbf{0.81} & \textbf{0.82}/\textbf{0.82} \\
        DexiNed   & \gls*{hog}/\gls*{ssim}     & 0.60/0.62                   & 0.60/0.63                   & \textbf{1.00}/0.91 & 0.75/0.74                   & 0.48/0.54                   \\
        Canny     & \gls*{msssim}/\gls*{ssim}  & 0.60/0.60                   & 0.62/0.61                   & 0.91/0.91          & 0.74/0.73                   & 0.52/0.52                   \\
        Laplacian & \gls*{lpips}/\gls*{lpips}  & 0.65/0.68                   & 0.71/0.71                   & 0.72/0.80          & 0.71/0.75                   & 0.64/0.65                   \\
        Mean Ch.  & \gls*{lpips}/\gls*{msssim} & 0.63/0.65                   & 0.62/0.65                   & 0.99/\textbf{0.94} & 0.76/0.77                   & 0.53/0.58                   \\
        \bottomrule
    \end{tabular}
    \caption{
        Quantitative performance analysis of the tampering detection using a decision tree with depth one.
        The metric indicates the selection for thresholding during the training of the decision tree.
        We report metric names and scores for \textit{predicted / ground truth} keypoints.
    }
    \label{table:eval:tamp:simple}
\end{table*}

Results in \cref{table:eval:tamp:simple} using predicted keypoints show that \meme{SimSaC}{\gls*{lpips}} yields the best performance and reaches $0.81$ accuracy and an F1-Score of $0.83$.
The by far highest precision is also achieved by \meme{SimSaC}{\gls*{lpips}}, which indicates cautious change detection for our use-case.
The highest recall is reached by \meme{DexiNed}{\gls*{hog}} and \meme{Mean Ch.}{\gls*{lpips}}, however, at the cost of precision.
Performance differences between using predicted and ground truth keypoint positions are comparatively small due to the high accuracy of the keypoint detection (\cf \cref{table:eval:aps}).

\subsubsection{Sensitivity Analysis: Tampering Types}
\label{sec:eval:tampering:types}

The analysis of performance differences across tampering types in \cref{table:eval:tamp:types}, shows that \textit{labels} can be detected most reliably, while \textit{tape} and especially \textit{writing (hard)} are more difficult to recognize. 
Surprisingly, when detecting \textit{writing} performance deteriorates when using ground truth keypoint annotations.
One potential reason for this might be, that inaccurate keypoints enlarge the region of interest unproportionally.

\begin{table*}[h!]
    \centering
    \begin{tabular}{lcccccc}
        \toprule
        Tampering ~~~~~~~           & \multicolumn{2}{c}{Label} & \multicolumn{2}{c}{Tape} & \multicolumn{2}{c}{Writing}                                  \\ 
        Type                        & ~~easy~~                  & ~~hard~~                 & ~~easy~~                    & ~~hard~~ & ~~easy~~ & ~~hard~~ \\
        \midrule
        Number of Samples           & 606                       & 570                      & 462                         & 546      & 624      & 498      \\
        Recall (Pred. Keypoints) ~~ & 1.00                      & 1.00                     & 0.58                        & 0.48     & 0.87     & 0.52     \\
        Recall (GT Keypoints)       & 1.00                      & 0.99                     & 0.59                        & 0.49     & 0.80     & 0.36     \\

        \bottomrule
    \end{tabular}
    \caption{
        Sensitivity analysis on the performance differences across tampering types using \meme{SimSaC}{\gls*{lpips}}.
    }
    \label{table:eval:tamp:types}
\end{table*}

\subsubsection{Sensitivity Analysis: Lens Distortion}
\label{sec:eval:tampering:lens}

\begin{figure}[h!] %
    \centering
    \includegraphics[width=.9\linewidth]{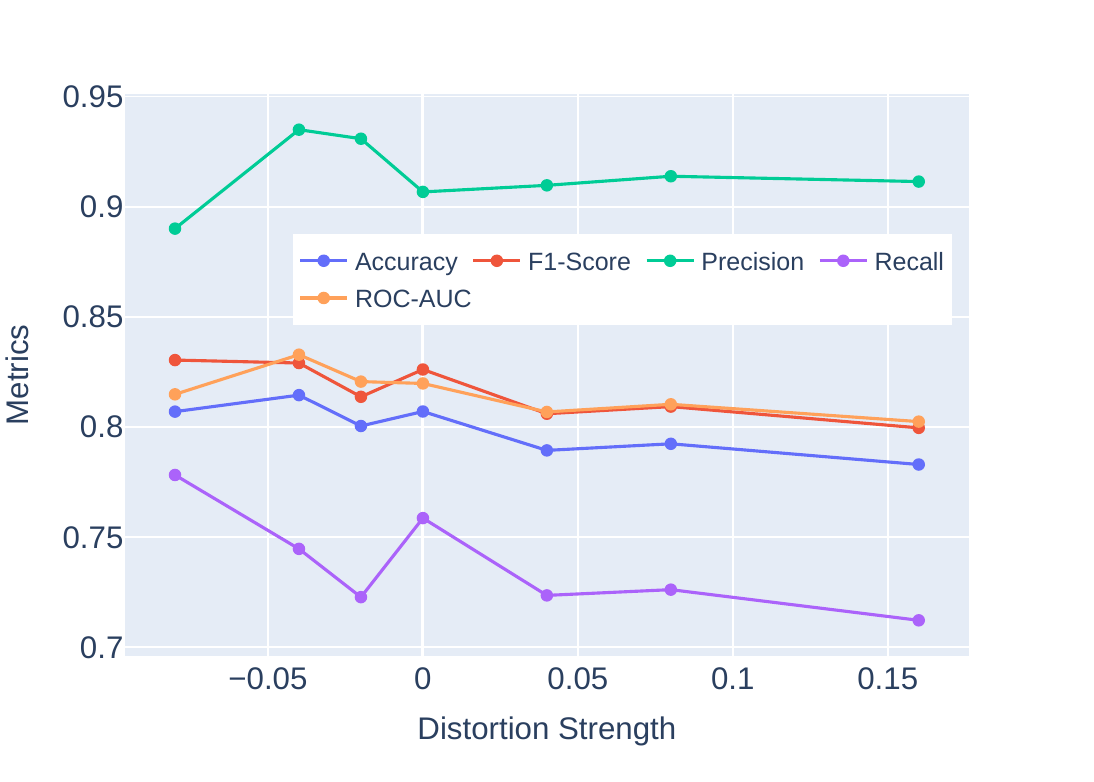}
    \caption{
        Sensitivity analysis for tampering detection \wrt to the distortion strength $A$ using pred. keypoints and \meme{SimSaC}{\gls*{lpips}}.
    }
    \label{fig:eval:tamp:distortion}
\end{figure}

We analyze the influence of six different degrees of distortion (\cf \cref{fig:eval:distortion:samples}) on the tampering detection quality using predicted keypoints and \meme{SimSaC}{\gls*{lpips}}.
These distortions imply that our simple perspective transformation cannot accurately create normalized side surface views and the change detection approach needs to handle these inaccuracies.
The results in \cref{fig:eval:tamp:distortion} suggest robustness \wrt distortions, with a slight negative trend for distortions with distortion strength $A>0$.
This is in line with the fact, that our approach can cope with lens distortion effects across the two real-world dataset \dstamp{} and \dsreal{}, while being trained on different, synthetic data.

\subsubsection{Sensitivity Analysis: Viewing Angles}
\label{sec:eval:tampering:angle}

\begin{figure}[h!] %
    \centering
    \includegraphics[width=.9\linewidth]{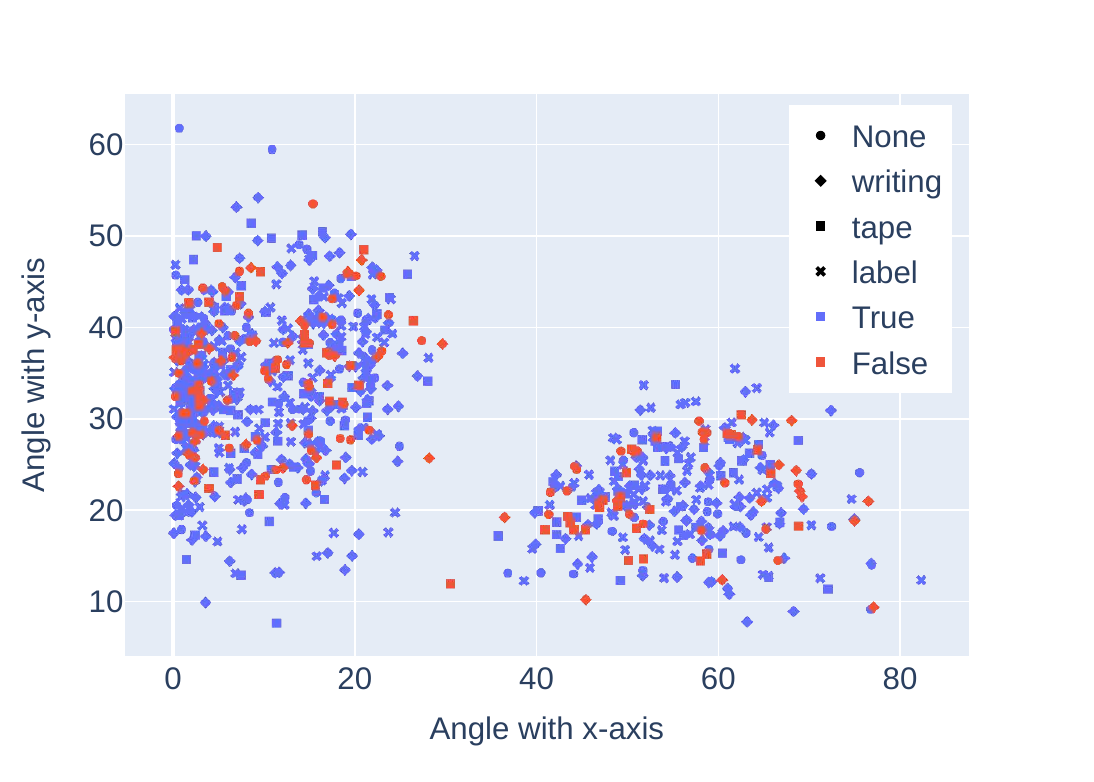}
    \caption{
        Sensitivity analysis for tampering detection \wrt to the viewing angle per side surface using predicted keypoints and \meme{SimSaC}{\gls*{lpips}}.
        Tampering types are encoded with different geometries and the prediction correctness with color-coding.
    }
    \label{fig:eval:angles}
\end{figure}

We approximate the viewing angle per parcel side surface by considering the angle between the x- and y-axis, and the polygon spanned by the four side surface corner points. 
No clear trend emerges from this analysis in \cref{fig:eval:angles}, which suggests that our approach is robust \wrt a reasonable spectrum of viewing angles. 
Note, however, that \dstamp{} does not feature extreme viewing angles.
Due to the strong distortions under such viewing angles, we expect the performance of tampering detection to degrade heavily. 

\section{Conclusion}
\label{sec:conclusion}

In this work, we propose a tampering detection pipeline for parcels that leverages keypoint and change detection.
We utilize the parcel keypoints to compute viewpoint-invariant parcel side surfaces, which reduces the problem of tampering detection to (1) identifying identical parcels and their side surfaces across time and (2) applying change detection on them.
We propose an unambiguous keypoint ordering for parcels and evaluate well-established baseline algorithms for the task of parcel corner point detection on real-world data.
Our approach reaches $75.76$ and $97.18$ \kpap{} on two real-world datasets when trained only on synthetic images from \dsparcel{} \cite{naumannParcel3DShapeReconstruction2023}.
Moreover, we introduce the first publicly available benchmark for tampering detection on parcels called \dstamp{}.
We propose a systematic approach for tampering detection which combines image homogenization approaches to alleviate lighting differences with several image similarity metrics.
To make a prediction, the most suitable metric is chosen and simple thresholding is applied.
Our approach performs best when combining SimSaC \cite{parkDualTaskLearning2022} with \gls*{lpips} \cite{zhangUnreasonableEffectivenessDeep2018} and reaches $0.81$ accuracy and an F1-Score of $0.83$.
The additional sensitivity analysis \wrt tampering types, lens distortion and viewing angles demonstrates the robustness of our approach.

Future work can incorporate recent advances in keypoint detectors and also exploit shape priors by utilizing a vanishing point loss \cite{ruiGeometryConstrainedCarRecognition2020} or enforcing 2D/3D consistency \cite{liRTM3DRealTimeMonocular2020}.
Furthermore, fine-tuning SimSaC \cite{parkDualTaskLearning2022} for tampering detection and more complex decision rules such as random forests are expected to improve performance significantly provided suitable datasets are available.
\printbibliography

\end{document}